\documentclass[conference]{IEEEtran}

\ifCLASSINFOpdf
\else
\fi
\usepackage{cite}
\usepackage{amsmath,amssymb,amsfonts}

\usepackage{algorithmic}
\usepackage{graphicx}
\usepackage{textcomp}
\usepackage{xcolor}
\usepackage{subfigure,bm,balance}
\usepackage[latin1]{inputenc}
\usepackage{amsmath}
\usepackage{amsfonts}
\usepackage{amssymb}
\usepackage{bbm}
\usepackage{dsfont}
\newcommand{\Tau}{\mathcal{F}}
\usepackage{graphics} 
\usepackage{booktabs}
\usepackage{float}
\usepackage[ruled,vlined]{algorithm2e}

\begin{document}
%
\title{Few Shot Learning Framework to Reduce Inter-observer Variability in Medical Images}
%
%
%
\author{\IEEEauthorblockN{Sohini Roychowdhury
}
\IEEEauthorblockA{Head of University Relations, Volvocars R\&D Tech Office USA, CA-94085,\\
}}
        
%
%

\maketitle

\begin{abstract}
Most computer aided pathology detection systems rely on large volumes of quality annotated data to aid diagnostics and follow up procedures. However, quality assuring large volumes of annotated medical image data can be subjective and expensive. In this work we present a novel standardization framework that implements three few-shot learning (FSL) models that can be iteratively trained by atmost 5 images per 3D stack to generate multiple regional proposals (RPs) per test image. These FSL models include a novel parallel echo state network (ParESN) framework and an augmented U-net model. Additionally, we propose a novel target label selection algorithm (TLSA) that measures relative agreeability between RPs and the manually annotated target labels to detect the ``best" quality annotation per image. Using the FSL models, our system achieves 0.28-0.64 Dice coefficient across vendor image stacks for intra-retinal cyst segmentation. Additionally, the TLSA is capable of automatically classifying high quality target labels from their noisy counterparts for 60-97\% of the images while ensuring manual supervision on remaining images. Also, the proposed framework with ParESN model minimizes manual annotation checking to 12-28\% of the total number of images. The TLSA metrics further provide confidence scores for the automated annotation quality assurance. Thus, the proposed framework is flexible to extensions for quality image annotation curation of other image stacks as well. 
\end{abstract}


%
\IEEEpeerreviewmaketitle

\section{Introduction}
Few shot learning (FSL) is a machine learning technique that is motivated by the human behavior of leveraging from strong knowledge priors and applying the contextual information from a small data set to new unseen scenarios \cite{fewshot1}. While FSL for object based classification from images has been well studied \cite{fewshot2}, few frameworks have been analyzed for the medical image segmentation domain till date \cite{fewshot1} \cite{mondal}. Further, transfer learning has limited scope in the medical image domain \cite{transferlearning}, which necessitates ``learning from less data". In this work we implement FSL methods to learn and generalize from small batches of optical coherence tomography (OCT) images acquired from a wide variety of vendors that typically represent 10-15$\mu m$ axial resolution.

Some recent works have demonstrated FSL frameworks by implementing encoder-decoder networks that can scale across different anatomical regions of interest (ROIs) on volumetric CT scans \cite{fewshot1}. However, there have been no such standardization efforts for OCT images acquired by a variety of vendors despite the evidence of a significant image quality variations for such high resolution image stacks \cite{sanchez}. Typically, the process of cataloging quality annotated data involves two steps. First, image batches are annotated by multiple graders, followed by the second tedious step in which a portion of randomly or carefully sub-sampled annotations are subjected to manual quality checking. Thus, there is a need for reliable, reproducible and repeatable decision making systems that can automate the quality assurance process, specifically for medical images. 

The OPTIMA cyst segmentation challenge (OCSC) \cite{optima} data set analyzed in this work contains OCT image stacks acquired by a variety of imaging vendors, where, each image is manually annotated by two manual graders/annotators, such that the dice coefficient (DC) between the graders ranges from $0.68-0.88$ \cite{sanchez}. Although ``aggregated" DCs reported across vendors in \cite{girishnew} \cite{Gopinath} and \cite{sanchez} are shown to have minimal variations across manual graders, the analysis per vendor image stack illustrates high variabilities across manual graders. For instance, the multi-scale CNN based model in \cite{sanchez} reports aggregated DCs of 0.56, 0.55, 0.54 across all vendor stacks for target labels (TLs)/graders . However, detailed analysis of most vs. least agreeable labellings across vendor stacks has DC of $[0.76, {\bf 0.78}, 0.71/ {\bf 0.65}, 0.63, 0.59]$ with respect to annotations for $G_1, G_2, G_1 \cap G_2$, on the [Spectralis/Topcon] stacks, respectively.
This example shows that both annotators $G_1$ and $G_2$ either over annotate or under annotate for certain image stacks, thereby having similar aggregate level annotations but highly dissimilar per-image level annotations. This motivates the need for per-image level standardization/selection of the best annotation as groundtruth.

\begin{figure}[ht!]
    \centering
    \includegraphics[width=0.45\textwidth]{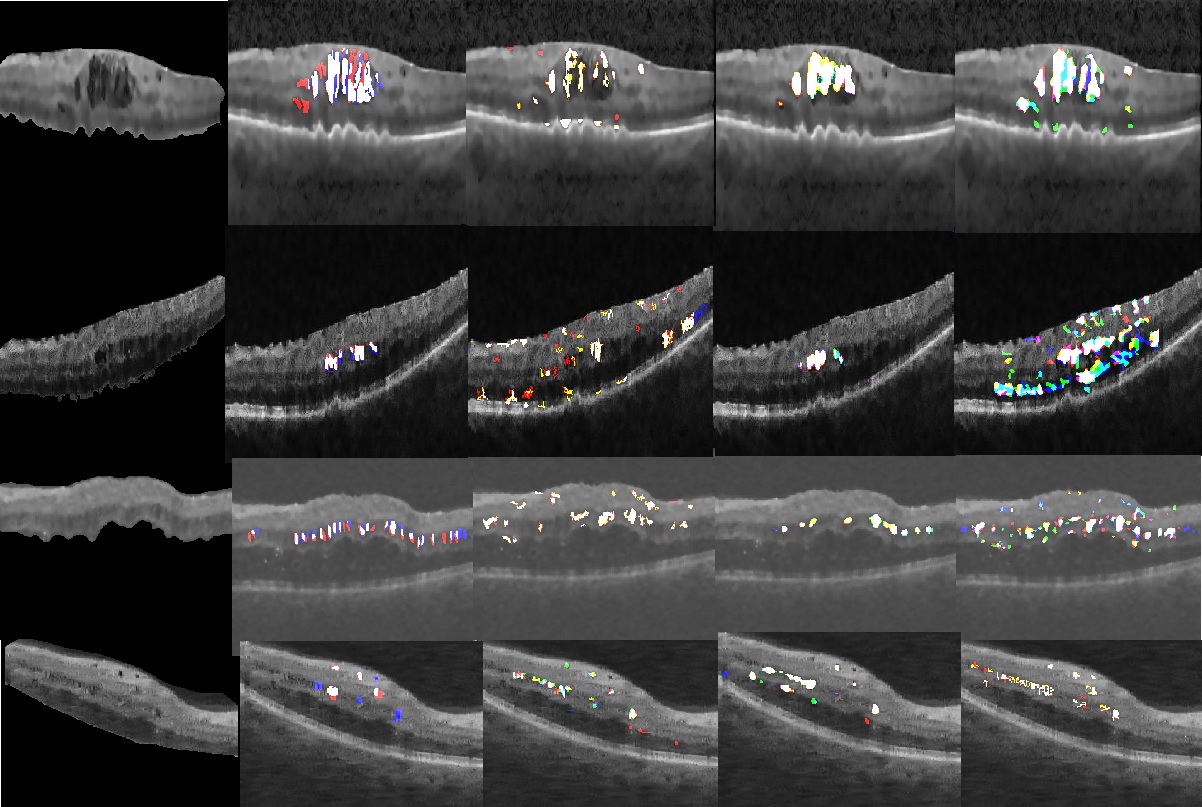}
    \caption{Examples of FSL models for cyst segmentation on Spectralis (Row 1), Topcon (Row 2), Cirrus (Row 3) images, Nidek (Row 4), respectively. Column 1: Original image with masked ROI, Column 2: TLs, Red: $G_1$, Blue: $G_2$, White: $G_1 \cap G_2$. Columns 3,4,5: Show three RPs represented as red, green and blue planes, respectively, on Baseline, U-net and ParESN methods, respectively.}
    \label{ex}
\end{figure}

This paper makes three major contributions. First, we present a novel system setup where three regional proposals (RPs), corresponding to segmented cysts, are predicted per image using three FSL models. Images with highly agreeable RPs (high DC) are relatively simple image annotation use cases, while low RP DC generally indicates unforeseen difficulty at image level for annotation. The three variants of FSL models are trained on atmost 5 images per vendor stack, and tested on the remaining images per stack. The FSL methods include a baseline that involves global thresholding, a novel Parallel Echo State Network (ParESN) model and the U-net model \cite{GIRISH}. Examples of RPs per vendor stack versus the manually annotated TLs are shown in Fig. \ref{ex}.
Second, we introduce a noisy target label generation function to verify the performance of the proposed novel rule-based \textit{Target Label Selection Algorithm} (TLSA). We observe that for 60-97\% of all images, the higher quality TL gets automatically selected while ensuring manual intervention for most of the remaining images. Third, we analyze the performance of the TLSA to automatically detect the ``best" annotation per image to reduce the volume of images subjected to manual quality assurance. The TLSA is capable of reducing manual quality checking to 12-28\%  and 4-46\% of total images using ParESN and U-net models, respectively . Additionally, the TSLA metrics provide a confidence score for the automated annotation quality assurance process, thereby standardizing the qualitative annotation process.

\section{Related Work}\label{rel}
Intra-retinal cyst segmentation plays a key role in timely diagnosis and treatment of diabetic macular edema, diabetic retinopathy, ocular inflammation, retinal vein occlusion and age-related macular edema \cite{GIRISH} \cite{Gopinath}. Till date several 3D image stack filtering and deep learning methods have been implemented on the OCSC dataset. The image-filtering methods in \cite{wilkins},\cite{Gopinath}, \cite{girishold}, \cite{esmarelli} and \cite{oguz} involve similar system setup that includes speckle noise removal followed by isolation of the ROI between the inner limiting membrane (ILM) and retinal pigment epithelium (RPE) layers. The next steps involve 2D-3D image filtering followed by thresholding and post-processing to eliminate false positive regions from cysts. For instance, \cite{wilkins} invokes bilateral filtering to achieve 0.43 DC for Spectralis images, while \cite{girishold} used watershed transform to achieve 0.7 DC for the same OCT vendor stack. The filtering method in \cite{Gopinath} uses active contour and complex diffusion methods to achieve 0.53-0.66 DC across all the 4 vendor stacks in OCSC data set. The work in \cite{esmarelli} uses curvelet k-SVD for volume level segmentation on Spectralis images only to achieve DC ranges of (0.45-0.71), (0.46-0.72), (0.45-0.73) against annotations from $G_1, G_2$ and $G_1 \cap G_2$, respectively, thereby indicating variance at image stack level in annotations. The baseline global thresholding FSL model in this work is inspired by these image filtering works.

The deep learning models for cyst segmentation on the other hand involve training from 1400-1600 B-scans across vendor image stacks followed by heavy data augmentation by either image flipping, width, height, zoom shifts \cite{GIRISH} \cite{girishnew} or by sub-sampling the images into 4.4 million smaller [21x21],[41x41] and [81x81] sub-images in \cite{sanchez}. Further, \cite{sanchez} shows test DC on Spectralis, Topcon, Nidek and Cirrus vendor image stacks are [0.64,0.51,0.28,0.42],respectively. This demonstrates the variabilities in image quality across vendors, thereby motivating the FSL methods that train on small data batches and apply the knowledge to new vendor image stacks, as presented in this work. The U-net models explored in \cite{GIRISH} and \cite{Gopinath} have comparable performance to \cite{sanchez} with DC of 0.56 across vendors. This motivates our use of the U-net model with data augmentation. 

\section{Mathematical Framework}\label{maths}
In this work, we train an automated standardization framework that utilizes FSL models and small batches of training data to predict multiple RPs ($\Pi$) for the object of interest in each test image. The relative overlaps between the RPs and the annotation TLs ($T$) are further analyzed using a novel rule-based method (TLSA) to generate a decision $\mathbf{\tau}$ per test image as shown in \eqref{eq0}. 

\begin{align}\label{eq0}
    \mathbf{\tau}=\Tau(I,\Pi,T), \text{where, } \tau=\{0,G_1,G_2.\dots\}\\ \nonumber
    \Pi=\{P_1,P_2,P_3\dots\}, 
    T=\{G_1,G_2,\dots \}.
\end{align}

Each OCT image ($I$) under analysis is resized to dimension $d= [300 \times 300]$ (motivated by \cite{Gopinath}) and subjected to pre-processing to extract a bottom-hat transformed version ($I_b$) (Section \ref{baseline}), image gradient magnitude and direction enhanced versions as $I_g$ and $I_d$, respectively. Thus, for each image $I$, $n=4$ input planes are utilized for the FSL models. For the baseline method that involves global thresholding, the input matrix $\mathbf{U}=I_b \in \mathbb{R}^{d}$. The parameters that need to be estimated for this method include: circular structure element diameter ($s_d$) and optimal threshold value ($\theta$) using a grid search approach. The output RPs are a result of slight variations in the estimated threshold as shown in \eqref{threshold}.

\begin{align}\label{threshold}
\forall I_b=g(I,s_d), P_1=I_b>\theta-0.05, P_2=I_b>\theta,\\ \nonumber
P_3=I_b>\theta+0.05.
\end{align}
As a second FSL method, a novel \textit{ParESN} model is proposed, such that each input image plane is resized to a single row ($d'=[1 \times 90,000]$), thereby resulting in an input matrix $\mathbf{U}\in \mathbb{R}^{n\times d'}$ as shown in \cite{ESNSRC}. Three independent reservoir layers with $m>n$ sparse connections are randomly initialized with weights $W_{\nu}$ for $\nu={1,2,3}$, respectively, and spectral radius adjustments \cite{ESNSRC}. Also the input weights to each reservoir are randomly assigned as $W_{in,\nu}$. The reservoir states for all pixels in the image $I$ are represented by matrix $\mathbf{X}=\{x(k), \forall k=1:d'\} \in \mathbb{R}^{m\times d'}$. The state update rule in \eqref{eq1} is applied in the training stage, which implies that the reservoir state in the same pixel position from the previous image ($x(k-1)$) affects the current image pixel ($x(k)$) (See supplementary material).
\begin{figure}[ht!]
    \centering
    \includegraphics[width=3.2in, height=2.0in]{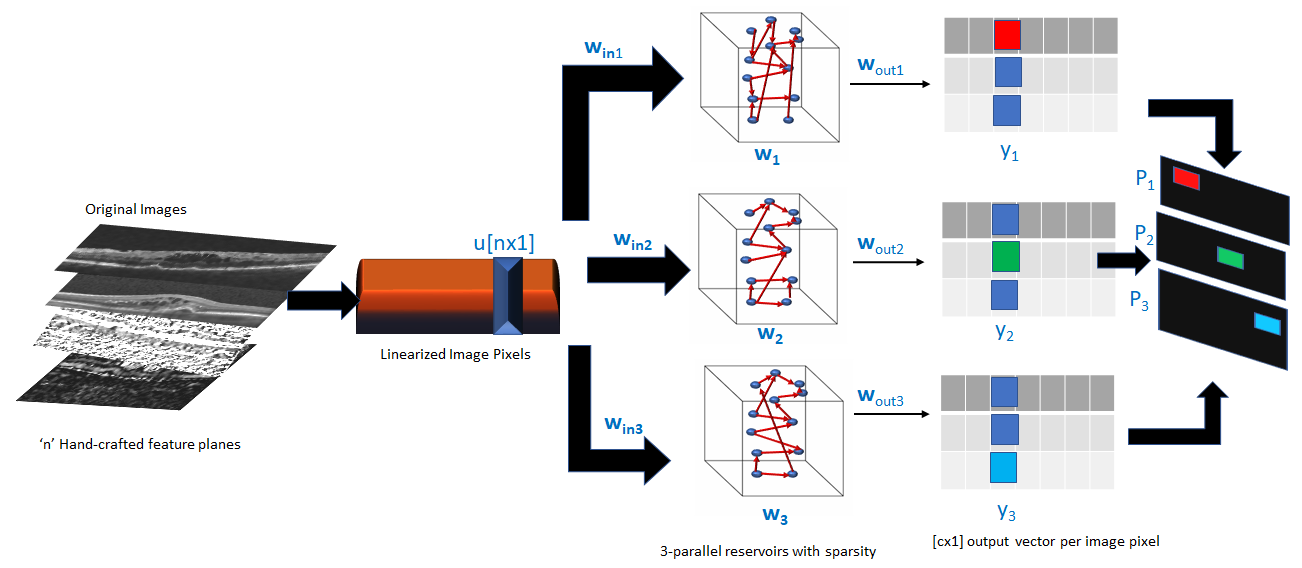}
    \caption{System setup of the proposed ParESN model.}
    \label{esn}
\end{figure}
\begin{align}\label{eq1}
\mathbf{x_{\nu}}(k)=(1-\alpha) \,\mathbf{x_{\nu}}(k-1)\\ \nonumber
+\alpha \,f\big(\mathbf{W}_{\mathrm{in, \nu}} \,[1;\mathbf{u}(k)]+\mathbf{W_{\nu}}\,\mathbf{x_{\nu}}(k-1)\big).
\end{align}

At the end of training stage, $W_{out,\nu}$ are computed for each parallel layer using \eqref{eq2}. 
\begin{align}\label{eq2}
\mathbf{W}_{\mathrm{out,\nu}}=\Big(\sum_{l=1}^{L}\mathbf{z}_{\nu,l}(k)\mathbf{z}^{\mathrm{T}}_{\nu,l}(k)+\lambda \, \mathbf{\mathds{1}}\Big)^{-1}\Big(\sum_{l=1}^{L}\mathbf{z}_{\nu,l}(k)y(k)\Big)
\end{align}
where, $\mathbf{z_{\nu,l}}(k)=[1;\mathbf{u}(k);\mathbf{x_{\nu}}(k)]$ are the extended system states for the 3 parallel layers, evaluated over $l=\{1,2 \dots L\}$ training images, and $y(k)$ represents the target label at pixel location $k$.

For the third FSL method, the U-net model \cite{Gopinath} is implemented with input matrix $\mathbf{U}=I \in \mathbb{R}^{d}$. The training data set is heavily augmented along with minimization of the DC loss function in \eqref{eqdice}. 
\begin{align}\label{eqdice}
Loss=\sum_{k=1}^{d'}(1-\frac{2p(k)y(k)}{p(k)+y(k)+1})
\end{align}
where, $p(k)\in \Pi$ represent the RP pixels per epoch and $y(k) \in T$ are the TLs. The RPs are generated by applying dropout to certain layers as described in Section \ref{Unet}.

Once the RPs are generated, the next step involves evaluation of several regional overlap and pixel-based metrics that are used in combination by function $\mathcal(F)$ in \eqref{eq0} to identify either the most reliable TL, i.e., if $G_1$ is better than $G_2$ or vice versa, or invoke manual verification. In the notation below $\circ$ denotes element-wise multiplication of vectors or matrices. $\mathbb{I}_{(\Pi_{i'},\Pi_{i''})}$ represents the intersection over union (IOU)/Jaccard coefficient \cite{GIRISH} between the automated RPs $\forall (i',i'') \in \{1,2,3\}, i' \neq i''$ using \eqref{metrics:iou}. Additionally, $\mathbb{I}_{(\Pi_{i'},T_{j})}$ is the IOU between the $i'$th RP and the $j$th TL, i.e., if $T=\{G_1,G_2\}$ and $j=\{1,2\}$ using \eqref{metrics:iou}.

The variance over mean for the RPs ($\Phi_{\Pi}$) captures the relative dis-agreeability between each other in \eqref{metrics:Phip}. The mean IOU between the RPs and the $j$th target label can then be evaluated as $\mu_{\mathbb{I}_{(\Pi,T_j})}$ using \eqref{metrics:Phij}, such that a higher value implies a good target label $T_j$. Also, the RP with maximum overlap with the $j$th target label can be recorded as $\psi^{*}_{j}$ in \eqref{metrics:psi}. Finally, to capture the degree of pixel variation in the RPs with respect to the $j$ th target label, a sub-image $I'_{j}$ is computed in \eqref{metrics:I}, followed by the variance over mean ($\Phi_j$) in pixel intensities computed in \eqref{metrics:Phij}. The metrics $\Phi_{\Pi}, \Phi_{j}$ represent the degree of variance, hence high values imply bad proposals and target label, respectively.

\begin{align}
\mathbb{I}_{(A,B)}=\frac{A \circ B}{A+B-A \circ B} \label{metrics:iou}\\
\Phi_{\Pi}=\frac{\sigma^2_{\mathbb{I}_{(\Pi,\Pi})}}{\mu_{\mathbb{I}_{(\Pi,\Pi)}}}\label{metrics:Phip}\\
\psi^{*}_j=\arg\max_{i=1,2,3}\{\mathbb{I}_{(P_i,T_j)}\}\label{metrics:psi}\\
I'_{j}=I\circ (T_j\circ[(\sum_{i'=1,2,3}P_{i'})>0])\label{metrics:I}\\
\Phi_{j}=\frac{\sigma^2_{I'_{j}}}{\mu_{I'_{j}}}\label{metrics:Phij}.
\end{align}

\section{Materials and Methods}\label{meth}
\subsection{Intra-retinal Cyst Data}

The OCSC \cite{optima} data set under analysis in this work uses [3, 1, 3, 2] and [1, 1, 1, 1] image stacks from Spectralis, Nidek, Topcon and Cirrus vendors taken from the training and testing 1 data sets, respectively. Further, we select the images that have cysts in them (we eliminate images without annotations) \cite{Gopinath}. Each image under analysis represents intra-retinal cysts of varying sizes in pathological OCT image stacks annotated by two manual graders $G_1$ and $G_2$. Each image is subjected to speckle noise removal \cite{GIRISH} and image centering, based on \cite{wilkins}\cite{Gopinath}, to center the image between ILM to RPE layers. To train the ParESN and U-net models, we use the first 3 images per vendor stack and 2 images at the mid point of the OCT stack. All remaining images are used for testing. Thus, the numbers of training/test images for the Spectralis, Nidek, Topcon, Cirrus data sets are [20,10,20,15/32,105,207,236], respectively.

\subsection{Image Pre-processing}
\begin{figure}[ht!]
    \centering
    \includegraphics[width=3.2 in, height=2.0 in]{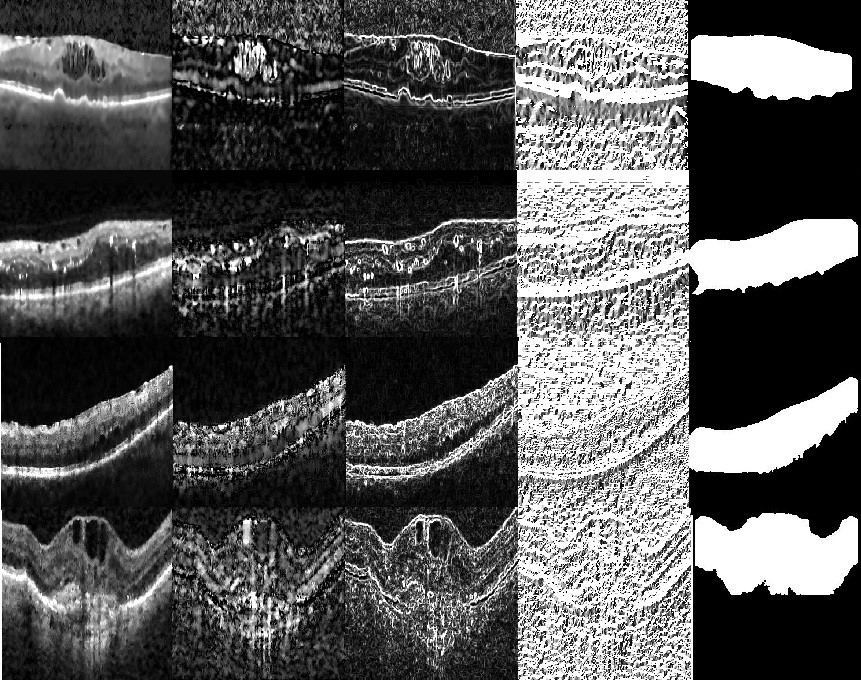}
    \caption{Pre-processed OCT image planes from multiple vendors. Row 1: Spectralis, Row 2: Topcon, Row 3: Nidek, Row 4: Cirrus. Column 1: Image $I$. Column 2: Bottom hat transformed image. Column 3: gradient magnitude. Column 4: gradient direction. Column 5: ROI mask.}
    \label{prep}
\end{figure}

Image planes that are processed with domain knowledge can be significantly useful for semantic segmentation tasks from medical images \cite{esmarelli} \cite{wilkins}. The OCT images analyzed here are single plane gray-scale representations of the intra-retinal thicknesses. The semantic segmentation effort here is directed towards isolating the cystic regions that appear as dark and hollow regions between the ILM and RPE layers. With this domain knowledge regarding cyst appearance, bottom-hat transformation \cite{bothat} can be applied to the images to highlight the cystic/hollow ROIs. Additionally, the cystic boundaries and gradient in the cystic regions with respect to the immediate surroundings are significant for cyst segmentation. Also, a mask corresponding to the intra-retinal regions between the ILM and RPE can be automatically generated by isolating the most significant edges from the top and bottom of the images, respectively, motivated by \cite{GIRISH} \cite{Gopinath}. Thus, the 4 pre-processed images shown in Fig. \ref{prep} become significant to our FSL models as they enhance the cystic characteristics. 

\subsection{Few Shot Learning Models}
The three FSL models under analysis are described below.
\subsubsection{Baseline, Global Thresholding}\label{baseline}
\begin{figure}[ht!]
    \centering
    \includegraphics[width=0.45\textwidth]{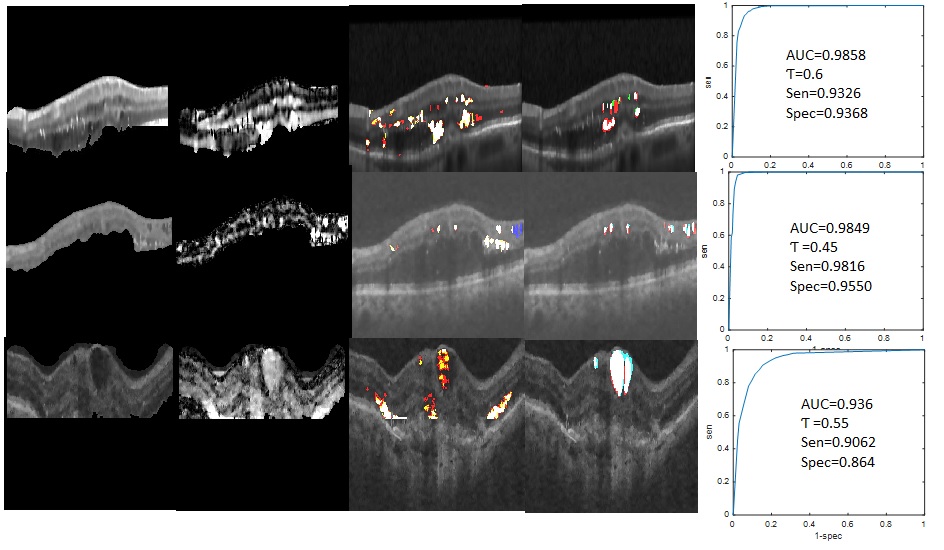}
    \caption{Example of the baseline global thresholding method on Spectralis (Row 1), Topcon (Row 2) and Cirrus (Row 3) images, respectively. Column 1: Original image with masked ROI, Column 2: Bottom hat filtered image with optimal $s_d$, Column 3: Thresholded predictions $P_1, P_2, P_3$ represented as red, green and blue planes, respectively, Column 4: annotated groundtruths $G_1, G_1 \cap G_2, G_2$ represented by red, green and blue planes, respectively, Column 5: best ROC with optimal parameters extracted from the image.}
    \label{base}
\end{figure}

In this method, a pre-processed image plane is subjected to a global threshold, that is empirically determined, to segment the cystic ROIs. Here, a training image is randomly selected from each vendor stack, followed by bottom-hat transformation to highlight the dark and hollow cystic regions. This operation involves finding several difference images between the closing and input image using a circular structuring elements of radius in range [$2:s_d$] with increments of 2. The maximum pixel value across all the transformations so far are retained as the bottom-hat transformed image \cite{bothat} followed by contrast adjustment in the range [0,1]. 
Next, global thresholding is performed with threshold values varying in the range [0,1] in increments of 0.05. Using a grid search for optimal $s_d$ in the range [3:25] in increments of 2, receiver operating characteristic curves (ROCs) \cite{bothat} are constructed for each vendor stack and the $s_d$ corresponding to ROC with highest area under ROC curve (AUC) is selected. Also, the threshold at the operating point (i.e. point on ROC that is closest to the left top corner) is recorded as $\theta$ followed by generation of the RPs ($P_1, P_2, P_3$), as shown in Fig. \ref{base}. The process of randomly selecting one training image, followed by optimal search of $s_d$ and $\theta$ and testing the performance on all other images from each image stack is performed 50 times per vendor stack and the average performance metrics are then analyzed. 

\subsubsection{ParESN Model with Masked Sub-images}
\begin{figure}[!ht]
	\centering
	\subfigure[Masked Spectralis image]
	{\includegraphics[height=1.6cm,width=0.2\textwidth]{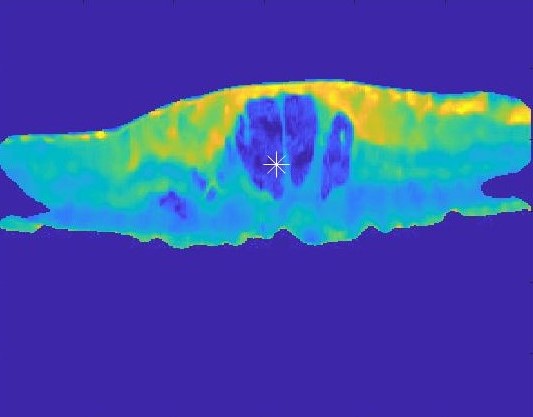}
		\label{c_ori}}
	\subfigure[Masks superimposed on image]
	{\includegraphics[height=1.6cm,width=0.2\textwidth]{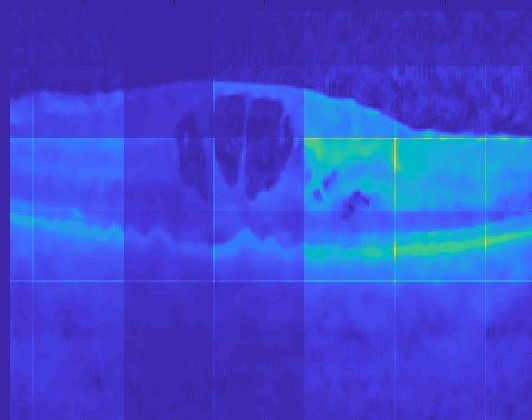}
		\label{c_gt}}
	\subfigure[Gradients in image]
	{\includegraphics[height=1.6cm,width=0.22\textwidth]{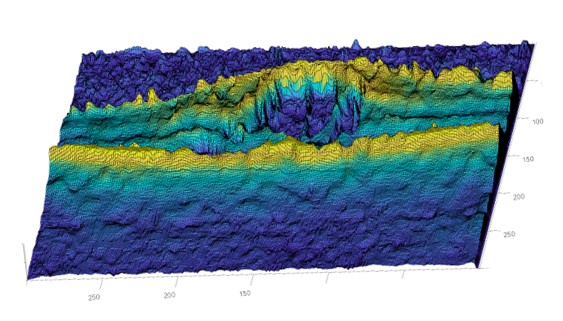}
		\label{fig:c_pred}}
	\subfigure[Gradients in masked image]
	{\includegraphics[height=1.6cm,width=0.22\textwidth]{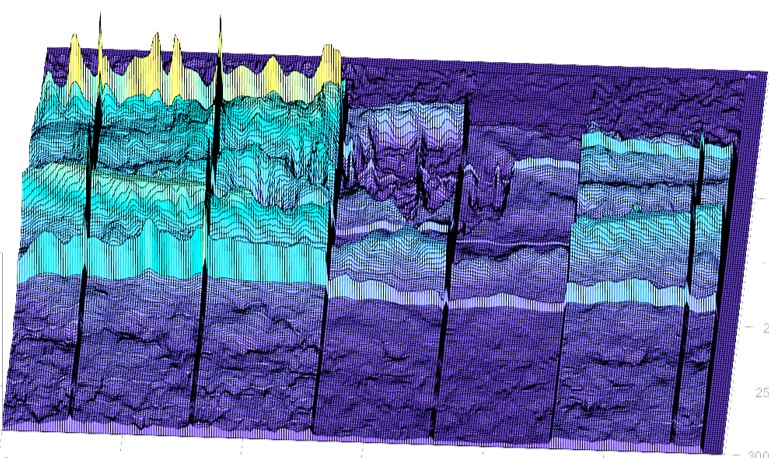}
		\label{fig:n_ori}}
	\caption{Qualitative importance of masked sub-images.}
	\label{outputscans}
	\vspace{-0.5cm}
\end{figure}

We implement a parallel ESN network model trained on masked sub-images generated per image. This model is motivated by the works in \cite{Gopinath} \cite{esmarelli} that propagate information across adjacent images in the 3D stack. Here, overlapping masks are extracted as follows by sub-sampling each image to represent variations in cyst sizes. First, centroid of the intra-retinal masked region is detected as shown in Fig. \ref{c_ori}. Next, starting from its centroid, masked regions of size $w_m=100$ are extracted as shown in Fig \ref{outputscans}(b). The corner masked regions are further resized to ensure $d=[100 \times 100]$. Each of the 4 pre-processed and masked image planes are then converted to input matrix $\mathbb(U)$ and fed as input to the ParESN model. The optimal reservoir size is empirically determined as $m=100$ by grid search \cite{ESNSRC}. 

\begin{figure}[ht]
    \centering
    \includegraphics[width=0.45\textwidth]{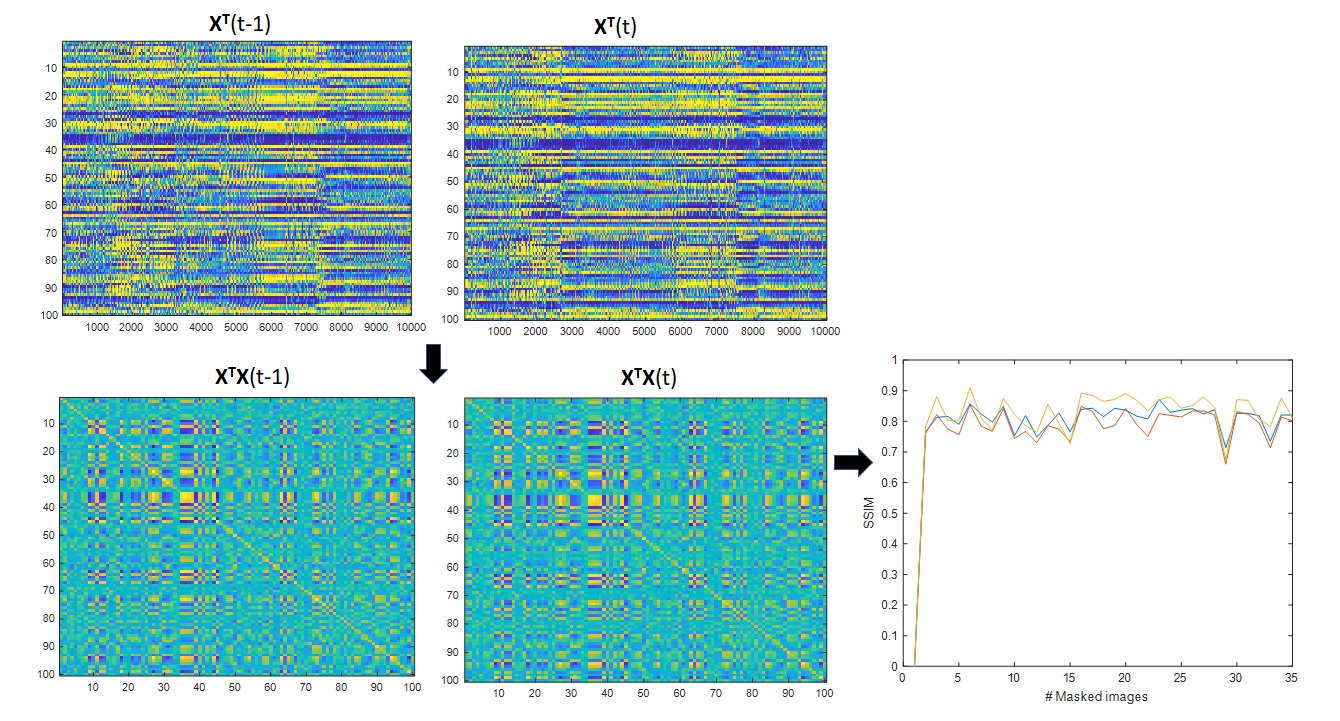}
    \caption{Image-based similarity metric for stopping criteria for \textit{ParESN} training.}
    \label{stop}
    	\vspace{-0.5cm}
\end{figure}

{\bf Stopping Criteria:}
The primary advantages of the ParESN model are low model parameter complexity and the presence of a stopping criteria to ensure model convergence in spite of being trained from a small training batch of images. The reservoir state matrices for each parallel layer $X \in \mathbb{R}^{m \times d'}$ for consecutive training epochs are shown in Fig. \ref{stop}. As a stopping criteria, we utilize the structural similarity index metric \cite{ssim} between $X^{T}X$ in consecutive sub-masked images epochs. We observe that as the number of masked sub-images subjected to training increase, the SSIM for $X^{T}X$ stabilizes with mean of 0.8 or more and standard deviation of 0.1 or less for at most 5 images, which is referred to the stopping criteria for training. Also, the use of 5 training images aligns with prior work in \cite{mondal}.

\subsubsection{Deep Learning, U-net Model}\label{Unet}
The U-net encoder-decoder network has been widely used for semantic segmentation tasks on medical images \cite{Gopinath} \cite{GIRISH}. In this work, we implement a 4 layer u-net model in \cite{unet} that is trained using the 5 training images (since 5 images used to train ParESN model) resized to [256x256] from each vendor image stack along with significant data augmentation that includes random horizontal flipping, rotation (range 20), width, height and zoom shifts in the range of 0.1 \cite{GIRISH} \cite{unet}. To ensure convergence, batch normalization is performed after every convolution step in the encoding phase with learning rate of $10^{-5}$ and DC as the loss function in \eqref{eqdice}. Finally, three RPs are generated for each test image by applying dropout of 0.5 for training and test images at the end of the third and fourth encoder convolutional layers. The dropout operation is not performed at the first two layers to ensure structural features being appropriately extracted. The training starts from the first Spectralis image stack and follows such that for every subsequent vendor image stack, the weights learnt from the previous image stack are initialized (transfer learning) and the training process continues till there is no further improvement in DC loss per vendor stack \eqref{eqdice} or upto 20 epochs are reached.

\subsection{Optimal Annotation/TL Selection}
Once the RPs are generated per test image, the next step is identifying the best annotation from a set of manually annotated TLs. However, for most publicly available medical data sets involving 2 manual annotations, there is no ground-truth to indicate the optimal TL per image, i.e., there is no method to verify if $G_1$ is better than $G_2$ for each labelled image or vice versa. Hence, our first step in quantitative evaluation of the TSLA ($\mathcal{F}$) is to generate varying degrees of noisy TLs, while evaluating the performance of $\mathcal{F}$ to select the actual TL over the noisy target. 

\begin{figure}[ht!]
    \centering
    \includegraphics[width=3.3in, height=1.0in]{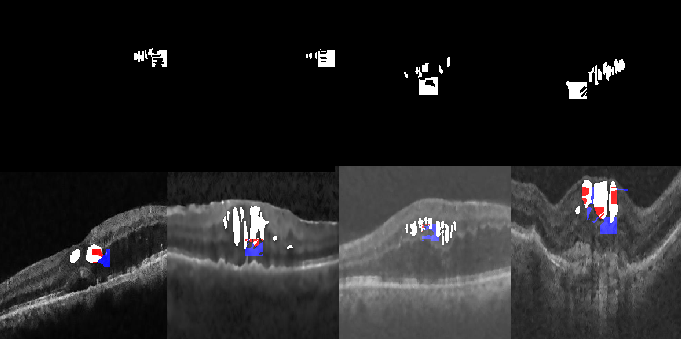}
    \caption{Examples of the noisy TL generation using the RCAP function. Row 1: Modifications made to TL $G_1$. Row 2: Noisy labels represented in blue image plane.}
    \label{sel}
\end{figure}

For this purpose we implement a random crop and paste (RCAP) function, which takes as input the TL whose noisy version is to be generated ($T$), a sub-window size ($w<100$) and some random seed pixel locations ($s_r,s_c$) corresponding to the locations where $T=1$. The output of this function is the noisy targets ($G_{1,n}, G_{2,n}$, respectively). Each iteration of the RCAP function first generates a random direction indicator in the range 1 to 8. The goal is to crop a square windowed region from target image $T$ (with side width $w$) that contains the seed pixel locations and rotate this windowed region based on the direction indicator function. Finally a flag (value 0 or 1) is randomly generated. If flag$=0$, or flag$=1$, the rotated version of the windowed sub-image or its negative (1-window pixels) is pasted back to the target image, respectively. The RCAP function is invoked iteratively from 1 to $\kappa=4$ times to demonstrate increasing degrees of noisy TLs. Examples of varying degrees of added noise to TL $G_1$ are shown in Fig \ref{sel}.

The final step is the TLSA ($\mathcal{F}$) shown in Algorithm \ref{algo1}{\footnote{Implementation Details: https://github.com/sohiniroych/Paralllel-ESN-For-Image-Quality}} to generate a decision per test image. Prior to the algorithm, the TLs are verified for individual quality. If both TLs are blank images (no annotations, $\sum_{k}T_{j}(k)=0$) then manual intervention is needed ($\tau=0$). If either of the target labels has some annotated regions, then the Algorithm 1 is initiated, where checks are performed for significant variation between TLs $T_1$ and $T_2$. Thus, $\eta=1$ implies very small variations (less than $w$ pixels) and $\eta=0$ implies significant variation, respectively. If there is a small variation between TLs, and the RPs vary significantly (large $\Phi_{\Pi}$), that implies the variations in TLs are too small for the automated system to detect, thereby requiring manual attention ($\tau=0$ returned). However, if the TLs vary significantly ($\eta=0$), then the metrics from \eqref{metrics:Phip}-\eqref{metrics:Phij} are computed. If the mean overlap of a particular TL with the RPs is significantly greater than the other TL (Ratio$_{\mu}$), then the larger overlapping TL is returned as best. Conversely the TL with least variance over mean pixel coefficient (Ratio$_v$) is the better TL. However, if (Ratio$_{\mu}$,Ratio$_{v}$) are both around unity (implying relatively small variability between target to RPs), and if the particular RP that has maximal overlap with TLs ($\Psi^{*}$) is similar across the TLs, then the Ratio$_\mu$ serves as decisive factor to return the best TL. However, if the TLs overlap with different RPs, i.e., $\psi^{*}_1=P_2$, $\psi^{*}_2=P_3$, this would imply difficulty in discerning between labels, in which case manual intervention will be needed ($\tau=0$ returned). Here, the thresholds are empirically selected as $\{\delta_1=0.4, \delta_2=1.1, \delta_3=1.02\}$. The metrics [Ratio$_{\mu}$,Ratio$_{v}$] serve as a confidence score for the evaluation process, such that higher metrics would imply higher confidence of discerning between TLs.
    \begin{algorithm}
    \caption{Target Label Selection Algorithm ($\mathcal{F}$)}\label{algo1}
    \hspace*{\algorithmicindent} \textbf{Input: $\Pi=\{P_1,P_2,P_3\}$, $T=\{T_1,T_2\}$, $I$,$w$}\\
     \hspace*{\algorithmicindent} \textbf{Output: $\tau$} 
    \begin{algorithmic}[1]
    \IF{$(\sum_{k}(T_1\circ T_2)-\min(\sum_{k}T_1,\sum_k T_2))<w$}
    \STATE $\eta \leftarrow 1$
    \ELSE
    \STATE $\eta \leftarrow 0$
    \ENDIF
    \STATE Compute [$\Phi_{\Pi}$]
    \IF{$\eta=1$ and $\Phi_{\Pi}>\delta_1$}
    \STATE  $\tau \leftarrow 0$
    \ELSE
        \STATE Compute [$\mu_{\mathbb{I}_{(\Pi,T_j)}}, \Psi^{*}_{j}, \Phi_{j}$]
        \IF{$\mu_{\mathbb{I}_{(\Pi,T_j)}} = 0 $}
            \STATE $\tau \leftarrow 0$
        \ENDIF    
        \STATE Ratio$_{\mu} \leftarrow \frac{\mu_{\mathbb{I}_{(\Pi,T_1)}}}{\mu_{\mathbb{I}_{(\Pi,T_2)}}}$
        \STATE Ratio$_{v} \leftarrow \frac{\Phi_2}{\Phi_1}$
        \IF{(Ratio$_{\mu} > \delta_2$) Or (Ratio$_v >\delta_3$)}
            \STATE $\tau\leftarrow 1$
        \ELSIF{(Ratio$_{\mu} < \frac{1}{\delta_2}$) Or (Ratio$_v <\frac{1}{\delta_3}$)}
            \STATE $\tau\leftarrow 2$
        \ELSIF{$\Psi^{*}_1=\Psi^{*}_2$}
            \STATE $\tau \leftarrow \arg\max_{j=1,2} ($Ratio$_{\mu},\frac{1}{\text{Ratio}_{\mu}})$
        \ELSE
            \STATE $\tau \leftarrow 0$
            
    \ENDIF
    \ENDIF
    \vspace{-0.1cm}
    \end{algorithmic}
    \end{algorithm}
    
\section{Experiments and Results}\label{expr}
In this work, we evaluate the proposed system that is trained on small batches of training images to generate RPs; whose relative dis-agreebility in terms of overlapping area with the TLs is indicative of standardized system level quality assurance. For this purpose we perform three major experiments. First, we analyze the performance of the three FSL models with respect to the different annotators for intra-retinal cyst segmentation in Section \ref{perform}. Second, we analyze the performance of the function $\mathcal{F}$ to automatically identify a TL from its noisy counterparts with varying degrees of additive noise generated using the RCAP function in Section \ref{perfRCAP}. Third, we utilize the function $\mathcal{F}$ to identify the ``best" TL from the two sets of manual annotations with the goal of reducing the number of images being subjected to manual quality assurance in Section \ref{perfFind}.

\subsection{Output Metrics}
The segmentation outputs from FSL models are evaluated at pixel level in terms of the cyst pixels that are correctly segmented (true positives, TP), pixels that appear in manual annotations but are missing in RPs (false negatives, FN), pixels that are mis-classified as cysts by the RPs (false positives, FP) and background pixels that are are detected as background (true, negatives, TN). DC is computed as $\frac{2TP}{2TP+FP+FN}$, Jaccard/IOU is $\frac{TP}{TP+FP+FN}$, sensitivity (sen) is $\frac{TP}{TP+FN}$, specificity (spec) is $\frac{TN}{TN+FP}$, accuracy (acc) is $\frac{TP+TN}{TP+FP+TN+FN}$. Performance analysis of the TLSA (accuracy) is in terms of the fraction of all test images that are classified to belong to a particular TL. 

\subsection{FSL Performance Analysis}\label{perform}
Table \ref{tab:datasets} presents the performance of the three FSL models implemented in this work for the test images ($L_t$). The average performance for RPs across vendor stacks is presented to assess the relative agreeability/dis-agreeability of the three FSL methods under analysis. The baseline global thresholding method has the least accuracy while the ParESN and U-net models have comparable segmentation performances. The best performance metrics for each vendor stack are highlighted in Table \ref{tab:datasets}. We observe that the RPs from ParESN have more variability than the U-net. Additionally, we observe that the U-net has better $spec$ metric while ParESN has better overall $sen$ metric. These observations reveal that the ParESN model is more likely to over-predict small cysts than miss them (low FN rate) while the U-net is more likely to miss small dispersed cysts and for the RPs to mostly agree (low FP rate).

Also, in Table \ref{tab:datasets}, we observe that for Spectralis, U-net and ParESN achieve greater than 0.55 DC while being trained on $G_1, G_2, G_1 \cap G_2$. This observation is aligned with \cite{Gopinath} that reports 0.56 DC for U-net. In \cite{sanchez}, a CNN model was trained on sub-sampled images from 2-3 vendor stacks and then tested on the remaining image stack, per vendor. We compare the test data set accuracy using $G_1$ between this work and our ParESN model. On the Spectralis, Nidek, Topcon and Cirrus image stacks, \cite{sanchez} and our ParESN achieved DC of (0.64,0.28,0.51,0.42) and (0.61,0.35,0.4,0.48), respectively. Thus, we observe that our FSL models trained on a few images achieve comparable to improved performances with respect to the deep learning counterparts that are trained on thousands of images.

 Finally, we observe that segmentation performances are significantly better for Spectralis and Cirrus vendor stacks when compared to Topcon and Nidek ones across all FSL methods. \cite{Gopinath} and \cite{sanchez} report similar trends across vendor stacks as well. For the Cirrus data set, the U-net outperforms ParESN since ParESN is unable to compensate for the motion blur in the image stacks as much as the U-net.
\begin{table}
   \centering
    \caption{Cyst segmentation Performances}\label{tab:datasets} 
    \scalebox{0.65}
    {
    \begin{tabular}{|c|c c c|c c c|c c c|}\hline
    \bf{Baseline}&&&$G_1$&&&$G_2$&&&$G_1 \cap G_2$ \\\hline
    RPs&$P_1$&$P_2$&$P_3$&$P_1$&$P_2$&$P_3$&$P_1$&$P_2$&$P_3$\\ \hline
    Spectralis&$L_t=32$&&&&&&&&\\ \hline
$DC$&0.3285&	0.3496&	0.3380&	0.3649&	0.3845&	0.3730&	0.3964&	0.4175&	0.4095\\
$IOU$&0.2127&	0.2287&	0.2204&	0.2577&	0.2728&	0.2644&	0.2872&	0.3034&0.2978\\
$sen$&0.4643&	0.4640&	0.4048&	0.4416&	0.4414&	0.3929&	0.5452&	0.5433&0.4868\\
$spec$&0.9577&0.9912&	0.9932&	0.9568&	0.9914&	0.9934&	0.9562&	0.9907&	0.9928\\
$acc$&0.9323&	0.9842&	0.9854&	0.9336&	0.9849&	0.9861&	0.9425&	0.9868&	0.9880\\\hline
Nidek&$L_t=105$&&&&&&&&\\ \hline
$DC$&0.2219&0.2484&	0.2255&	0.2718&	0.2988&	0.2786&	0.2956&	0.3230&	0.3079\\
$IOU$&0.1465&	0.1646&	0.1501&	0.1941&	0.2127&	0.2002&	0.2231&	0.2417&	0.2331\\
$sen$&0.3575&	0.3502&	0.2864&	0.3823&	0.3777&	0.3195&	0.4602&	0.4598&	0.3944\\
$spec$&0.9573&0.9917&	0.9939&	0.9579&	0.9918&	0.9939&	0.9568&	0.9915&	0.9937\\
$acc$&0.9198&0.9800&	0.9814&	0.9262&	0.9816&	0.9829&	0.9351&	0.9841&	0.9855\\\hline
Topcon&$L_t=207$&&&&&&&&\\ \hline
$DC$&0.2617&0.2850&	0.2670&	0.2594&	0.2760&	0.2604&	0.2696&	0.2939&	0.2836\\
$IOU$&0.1711&	0.1876&	0.1756&	0.1713&	0.1831&	0.1728&	0.1836&	0.2009&	0.1944\\
$sen$&0.3662&	0.3562&	0.2985&	0.3510&	0.3307&	0.2806&	0.4258&	0.4120&	0.3523\\
$spec$&0.9544&0.9925&	0.9945&	0.9566&	0.9926&	0.9945&	0.9529&	0.9920&	0.9940\\
$acc$&0.9222&	0.9813&	0.9826&	0.9227&	0.9802&	0.9814&	0.9299&	0.9835&	0.9850\\\hline
Cirrus&$L_t=236$&&&&&&&&\\ \hline
$DC$&0.3402&0.3588&	0.3400&	0.4358&	0.4536&	0.4298&	0.4678&	0.4864&	0.4681\\
$IOU$&0.2166&	0.2308&	0.2172&	0.3289&	0.3422&	0.3255&	0.3627&	0.3770&	0.3640\\
$sen$&0.3877&	0.3838&	0.3332&	0.4368&	0.4357&	0.3912&	0.5242&	0.5216&	0.4743\\
$spec$&{\bf0.9821}&{\bf0.9960}&{\bf	0.9972}&{\bf	0.9810}&{\bf	0.9956}&{\bf	0.9968}&{\bf	0.9794}&{\bf	0.9951}&{\bf	0.9964}\\
$acc$&0.9335&	0.9780&	0.9782&	0.9383&	0.9796&	0.9796&	0.9479&	0.9831&	0.9832\\ \hline
{\bf ParESN}&&&&&&&&&\\ \hline
Spectralis&$L_t=32$&&&&&&&&\\ \hline
$DC$&	{\bf0.6109}&{\bf	0.6033}&{\bf	0.5972}&{\bf	0.6042}&	{\bf0.5972}&{\bf	0.5936}&{\bf	0.6211}&{\bf	0.6124}&{\bf	0.6134}\\
$IOU$&	{\bf0.4496}&{\bf	0.4420}&{\bf	0.4386}&{\bf	0.4458}&	{\bf0.4383}&{\bf	0.4369}&{\bf	0.463}&{\bf	0.4590}&{\bf	0.4604}\\
$sen$&{\bf	0.6457}&{\bf	0.6439}&{\bf	0.6213}&{\bf	0.5906}&	{\bf0.5884}&{\bf	0.5711}&{\bf	0.7029}&{\bf	0.6993}&	{\bf0.6831}\\
$spec$&	0.9747&	0.9930&	0.9939&	0.9769&	0.9936&	0.9946&	0.9698&	0.9916&	0.9926\\
$acc$&	0.9584&	0.9887&	{\bf0.9896}&	0.9533&	0.9873&	0.9883&{\bf	0.9607}&{\bf	0.9892}&{\bf	0.9901}\\\hline
Nidek&$L_t=105$&&&&&&&&\\ \hline
$DC$&	{\bf0.3449}&{\bf	0.3348}&{\bf	0.3420}&{\bf	0.3767}&{\bf	0.3674}&{\bf	0.3718}&{\bf	0.4074}&{\bf	0.3991}&{\bf	0.4063}\\
$IOU$&	{\bf 0.2293}&{\bf 0.2220}&{\bf	0.2273}&{\bf	0.2625}&	{\bf0.2557}&{\bf	0.2590}&{\bf	0.3000}&{\bf	0.2940}&{\bf	0.2990}\\
$sen$&{\bf	0.6486}&{\bf	0.6481}&{\bf	0.6570}&{\bf	0.6690}&	{\bf0.6709}&{\bf	0.6764}&{\bf	0.7307}&{\bf	0.7365}&	{\bf0.7451}\\
$spec$&	0.9480&	0.9572&	0.9571&	0.9455&	0.9563&	0.9561&	0.8937&	0.9552&	0.9551\\
$acc$&	{\bf0.9504}&	0.9544&	0.9548&{\bf	0.9516}&	0.9548&	0.9548&	{\bf0.9536}&	0.9552&	0.9551\\\hline
Topcon&$L_t=105$&&&&&&&&\\ \hline
$DC$&	{\bf0.3974}&{\bf	0.3865}&{\bf	0.3952}&{\bf	0.3718}&{\bf	0.3743}&{\bf	0.3840}&{\bf	0.3666}&{\bf	0.3742}&{\bf	0.3859}\\
$IOU$&	{\bf0.2723}&{\bf	0.2636}&{\bf	0.2713}&{\bf	0.2546}&{\bf	0.2556}&{\bf	0.2641}&{\bf	0.2531}&{\bf	0.2580}&{\bf	0.2681}\\
$sen$&	{\bf0.5116}&{\bf	0.5079}&{\bf	0.5002}&{\bf	0.4851}&{\bf	0.4724}&{\bf	0.4695}&{\bf0.5692}&{\bf	0.5592}&{\bf	0.5547}\\
$spec$&	0.9719&	0.9924&	0.9932&	0.9668&	0.9923&	0.9931&	0.9640&	0.9914&	0.9923\\
$acc$&	{\bf0.9486}&{\bf	0.9843}&{\bf	0.9850}&{\bf	0.9417}&	{\bf0.9828}&{\bf	0.9835}&	0.9486&{\bf	0.9855}&{\bf	0.9864}\\\hline
Cirrus&$L_t=105$&&&&&&&&\\ \hline
$DC$&	0.4715&{\bf	0.4815}&	0.4610&	0.5736&	0.5812&	0.5668&	0.5846&	{\bf0.5898}&	0.5845\\
$IOU$&	0.3295&	0.3389&	0.3203&	0.4454&	0.4528&	0.4403&	0.4625&	0.4671&	0.4625\\
$sen$&	0.4604&	0.4842&	0.4321&	0.5483&	0.5649&	0.5299&	0.6252&	0.6480&	0.6081\\
$spec$&	0.9787&	0.9922&0.9935&0.9761&	0.9912&	0.9927&0.9735&0.9901&0.9918\\
$acc$&	0.9503&	0.9821&	0.9827&	0.9506&	0.9822&	0.9828&	{\bf0.9581}&{\bf	0.9846}&{\bf	0.9856}\\ \hline
{\bf U-net}&&&&&&&&&\\ \hline
Spectralis&$L_t=32$&&&&&&&&\\ \hline
$DC$	&0.5717& 0.5711&0.5655&	0.5759&	0.5768&	0.5707&	0.5574&	0.5558&	0.5519\\
$IOU$	&0.4314	&0.4308&	0.4251&	0.4337&	0.4343&	0.4282&	0.4172&	0.4158&	0.4116\\
$sen$	&   0.6078&	0.6155&	0.5942&	0.5762&	0.5842&	0.5647&	0.6418&	0.6495&	0.6284\\
$spec$&	{\bf 0.9757}&	{\bf 0.9937}&{\bf0.9942}&{\bf0.9789}&{\bf0.9947}&{\bf0.9951}&{\bf0.9724}&	{\bf0.9929}&{\bf0.9933}\\
$acc$	&   {\bf 0.9585}&{\bf0.9889}&	0.9888&{\bf	0.9573}&{\bf	0.9886}&{\bf0.9885}&	0.9595&	0.9891&	0.9892\\\hline
Nidek&$L_t=105$&&&&&&&&\\ \hline
$DC$ &  0.2379&	0.2366&	0.2425&	0.2850&	0.2858&	0.2947&	0.3279&	0.3250&	0.3366\\
$IOU$  & 0.1508&	0.1502&	0.1542&	0.1995&	0.2003&	0.2060&	0.2412&	0.2394&	0.2473\\
$sen$ &   0.2360&	0.2371&	0.2387&	0.2627&	0.2660&	0.2692&	0.3363&	0.3360&	0.3418\\
$spec$ &  {\bf 0.9920}&{\bf	0.9978}&{\bf	0.9978}&{\bf	0.9914}&	{\bf0.9976}&{\bf0.9977}&{\bf0.9904}&{\bf0.9973}&{\bf0.9974}\\
$acc$ &   0.9341&	{\bf 0.9809}&{\bf	0.9809}&	0.9423&{\bf	0.9832}&{\bf	0.9833}&	0.9515&{\bf	0.9858}&{\bf0.9859}\\\hline
Topcon&$L_t=105$&&&&&&&&\\ \hline
$DC$ &  0.3453&	0.3432&	0.3453&	0.3298&	0.3295&	0.3308&	0.3415&	0.3400&	0.3416\\
$IOU$  & 0.2348&	0.2333&	0.2351&	0.2212&	0.2210&	0.2222&	0.2336&	0.2325&	0.2339\\
$sen$ &   0.3679&	0.3660&	0.3689&	0.3440&	0.3439&	0.3467&	0.4209&	0.4196&	0.4229\\
$spec$ &  {\bf0.9853}&{\bf0.9962}&{\bf	0.9962}&{\bf	0.9851}&{\bf	0.9962}&{\bf	0.9961}&{\bf	0.9825}&{\bf	0.9954}&{\bf	0.9954}\\
$acc$ &   0.9443&	0.9832&	0.9832&	0.9406&	0.9820&	0.9820&	{\bf0.9514}&	0.9854&	0.9854\\\hline
Cirrus&$L_t=105$&&&&&&&&\\ \hline
$DC$ &  {\bf0.4802}&	0.4803&	{\bf0.4798}&{\bf	0.5810}&{\bf	0.5818}&	{\bf 0.5815}&{\bf	0.5890}&	0.5889&{\bf	0.5896}\\
$IOU$  &{\bf 0.3591}&{\bf	0.3590}&{\bf	0.3586}&{\bf	0.4673}&{\bf0.4677}&{\bf0.4672}&{\bf0.4788}&{\bf0.4783}&{\bf0.4788}\\
$sen$ &  {\bf 0.6376}&{\bf	0.6385}&{\bf	0.6351}&{\bf	0.6953}&{\bf	0.6973}&{\bf	0.6962}&{\bf	0.8109}&{\bf	0.8101}&{\bf	0.8106}\\
$$spec$$ &  0.9622&	0.9872&	0.9872&	0.9589&	0.9861&	0.9861&	0.9522&	0.9836&	0.9836\\
$acc$ &   {\bf0.9530}&{\bf	0.9832}&{\bf	0.9832}&{\bf	0.9545}&	{\bf0.9837}&{\bf	0.9837}&	0.9557&	0.9840&	0.9840\\ \hline
\end{tabular}
}
\vspace{-0.5cm}
\end{table}

\subsection{TL vs. Noisy TL Selection Analysis}\label{perfRCAP}
Having assessed the reliability of the FSL models, the next step is classification of actual TLs ($G_1, G_2$) from their noisy counterparts ($G_{1,n}, G_{2,n}$) produced by the RCAP function. In Table \ref{tab: selection}, the automated accuracy for the best TL selection for the different vendor image stacks is shown. The noisy TLs with increasing degrees of noise denoted by $\kappa=[1:4]$ are simulated 20 times per image and the average automated accuracy of selecting the actual TL as opposed to manual intervention is presented in Table \ref{tab: selection}. Here, an automated accuracy of 0.92 implies that for almost 8\% of the images in that stack, manual intention was required (i.e. $\tau=0$). Also, we observe that for all images from Spectralis and Cirrus stack, either the correct TL or manual intervention was selected. However, for the Nidek and Topcon vendor stacks, with low segmentation accuracy scores, 1.25-3.125\% of noisy TLs were selected for $\kappa=1$ over the actual TLs. These instances occurred when the TL modifications were too small, or when small cysts existed in the TL, or when additional pathologies were presented by the retinal image.

\begin{table}[ht]
   \centering
    \caption{Automated Accuracy for TL Selection. The mean (std deviation) of accuracy is presented below. Best metrics are highlighted.}\label{tab: selection} 
    \scalebox{0.85}
    {
    \begin{tabular}{|c|c c|c c|}\hline
    {\bf Target Labels}&$G_1$ vs $G_{1,n}$&&$G_2$ vs $G_{2,n}$&\\\hline
    Vendor Stack (Noise)&ParESN&Unet&ParESN&Unet\\\hline
Spectralis ($\kappa=1$)&	0.83(0.08)&	{\bf 0.85(0.11)}&	0.73(0.10)&	{\bf0.84	(0.10)}\\
$\kappa=2$&	0.91	(0.06)&	{\bf0.93	(0.05)}	&0.88	(0.07)&{\bf	0.94	(0.03)}\\
$\kappa=3$&	0.95	(0.04)&	{\bf 0.95	(0.02)}	&0.92	(0.06)&{\bf	0.95	(0.02)}\\
$\kappa=4$&	{\bf0.97	(0.03)}&	0.95	(0.02)	&0.95	(0.04)&{\bf	0.96	(0.01)}\\\hline
Topcon ($\kappa=1$)&	0.68	(0.04)&	{\bf 0.87	(0.12)}&{\bf	0.73	(0.04)}&	0.52	(0.04)\\
$\kappa=2$&	0.8	(0.03)&{\bf	0.94	(0.05)}&	{\bf0.84	(0.03)}&	0.61	(0.03)\\
$\kappa=3$&	0.84	(0.02)&{\bf	0.95	(0.02)}&{\bf	0.87	(0.03)}&	0.64	(0.02)\\
$\kappa=4$&	0.86	(0.02)&	{\bf0.96	(0.02)}&{\bf	0.88	(0.02)}&	0.65	(0.02)\\\hline
Nidek ($\kappa=1$)&	0.65	(0.04)&{\bf	0.85	(0.11)}&	0.59	(0.05)&	{\bf0.60	(0.04)}\\
$\kappa=2$&	0.74	(0.03)&	{\bf0.92	(0.04)}&	0.69	(0.04)&{\bf	0.73	(0.03)}\\
$\kappa=3$&	0.76	(0.03)&	{\bf0.93	(0.04)}&	0.70	(0.03)&	{\bf0.78	(0.03)}\\
$\kappa=4$&	0.78	(0.03)&	{\bf0.93	(0.03)}&	0.72	(0.03)&{\bf	0.80	(0.02)}\\\hline
Cirrus ($\kappa=1$)&	0.7	(0.05)&	{\bf 0.82	(0.10)}&	0.68	(0.07)&{\bf	0.82	(0.06)}\\
$\kappa=2$&	0.81	(0.04)&	{\bf0.88	(0.03)}&	0.85	(0.04)&	{\bf 0.91	(0.03)}\\
$\kappa=3$&	0.84	(0.03)&	{\bf 0.87	(0.04)}&	0.87	(0.03)&	{\bf 0.92	(0.01)}\\
$\kappa=4$&	0.85	(0.02)&	{\bf 0.88	(0.03)}&	0.90	(0.03)&{\bf	0.92	(0.01)}\\ \hline
 \end{tabular}}
    \vspace{-0.5cm}
\end{table}
From Table \ref{tab: selection}, we observe that as $\kappa$ increases, the accuracy of TL selection increases, which is an intuitive observation. For Spectralis vendor stack with high relative agreeability between RPs, TL $G_1$ and $G_2$ are equally reliable against their noisy counterparts. For Topcon, $G_1$ is more reliable using U-net model followed by $G_2$ using ParESN model against noisy labels, respectively. For Nidek, the RPs are least accurate and $G_1$ is more reliable against noisy labels than $G_2$. Finally, for Cirrus vendor stack, $G_1$ and $G_2$ have comparable performances against noisy labels.
\subsection{Optimal TL Selection}\label{perfFind}
\begin{figure}[ht!]
    \centering
    \includegraphics[width=3.3in, height=1.8in]{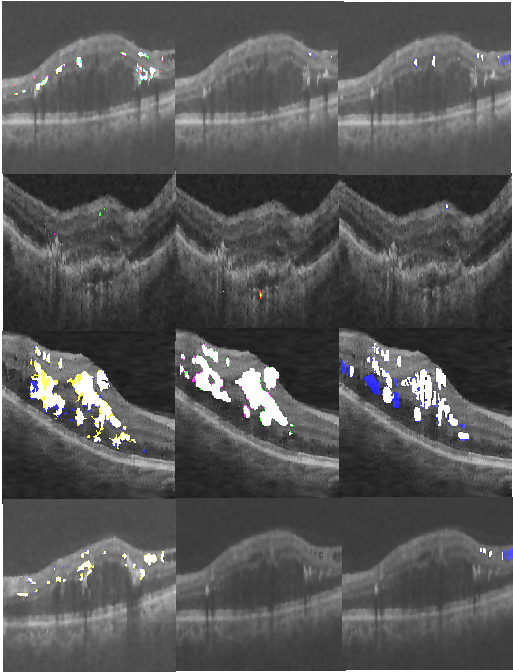}
    \caption{Examples of FSL mis-classifications followed by TLSA favoring the target labels from another annotator. Top and bottom rows correspond to Topcon and middle two rows are Cirrus images. Column 1: ParESN RPs trained on $G_1$, Column 2: U-net RPs trained on $G_1$, Column 3: TLs with $G_2=$blue, $G_1 \cap G_2=$white. In spite of mis-classified regions in the RPs, the TLs $G_2$ (blue plane) are preferred owing to high overall regional agree-ability between RPs and TLs.}
    \label{selection}
\end{figure}

Finally, the TLSA is applied on all vendor image stacks to automatically identify the best TL from $G_1, G_2$ and to minimize manual quality checking. The fraction of images in each vendor stack for which label ($\tau=1, G_1$) or ($\tau=2, G_2$) or manual intervention ($\tau=0$) is preferred are shown in Table \ref{tab:select}. Here, the following observations are made for each vendor.
\begin{itemize}
    \item Spectralis: Trained on $G_1$, FSL models favor $G_1$ more than $G_2$. Trained on $G_2$, there is equal affinity for $G_1$ and $G_2$ and trained on $G_1\cap G_2$ FSL models favor $G_1$ more than $G_2$.
    \item Topcon: Trained on $G_1$, ParESN model favors $G_1$ more than $G_2$. U-net model predicts several blank RPs (misses small cysts), thus requiring manual intervention on 40-46\% images. Trained on $G_2$ or $G_1\cap G_2$ both TLs $G_1$ and $G_2$ are equally likely as shown in Fig. \ref{selection}.
    \item Nidek: RPs are least reliable on this vendor stack. The FSL models favor the label the model is trained on. If trained on $G_1\cap G_2$, both $G_1$ and $G_2$ are equally preferable.
    \item Cirrus: The FSL models favor the TL used to train the FSL models. Trained on $G_1\cap G_2$, $G_1$ is more preferable than $G_2$ (similar to Spectralis).
\end{itemize}

\begin{table}[ht]
   \centering
    \caption{Fraction of Target Label Selection. Lower fraction of Manual intervention indicates a successful automated quality assurance process.}\label{tab:select} 
    \scalebox{0.85}
    {
    \begin{tabular}{|c|c c c|c c c|}\hline
    Model&&ParESN&&U-net\\ \hline
    Target label selected&$G_1$&$G_2$&Manual&$G_1$&$G_2$&Manual\\ \hline
     Spectralis&&&&&&\\ \hline
    Trained on $G_1$&	0.56&	0.19&	0.25&	0.62&	0.29&	0.08\\
 $G_2$&	0.35&	0.39&	0.25&	0.42&	0.41&	0.16\\
$G_1\cap G_2$&0.48&	0.25&	0.27&	0.64&	0.32&	0.04\\ \hline
Topcon&&&&&&\\ \hline						
Trained on $G_1$&	0.51&	0.28&	0.21&	0.31&	0.23&	0.46\\
$G_2$&	0.38&	0.38&	0.24&	0.28&	0.26&	0.46\\
$G_1\cap G_2$&	0.42&	0.42&	0.16&	0.36&	0.23&	0.40\\\hline
Nidek&&&&&&\\ \hline
Trained on $G_1$&	0.54&	0.28&	0.17&	0.51&	0.33&	0.15\\
$G_2$&	0.31&	0.56&	0.12&	0.36&	0.49&	0.14\\
$G_1\cap G_2$&	0.25&	0.48&	0.28&	0.56&	0.14&	0.29\\\hline
Cirrus&&&&&&\\ \hline						
Trained on $G_1$&	0.57&	0.20&	0.23&	0.68&	0.16&	0.16\\
$G_2$&	0.37&	0.46&	0.17&	0.29&	0.56&	0.15\\
$G_1\cap G_2$&	0.64&	0.08&	0.28&	0.51&	0.36&	0.13\\ \hline
    \end{tabular}}
    \vspace{-0.5cm}
\end{table}

\section{Conclusions and Discussion}\label{conc}
This work aims to standardize the quality of segmentation annotations per manual grader using a quantitative system-level framework. While existing works so far have considered inter-observer variability on an aggregate level in \cite{sanchez} \cite{Gopinath}, we demonstrate that such aggregation does not effectively mitigate inter-observer variability at a per-image level. The proposed system trains FSL models on small batches of data per vendor image stack to predict multiple RPs, followed by predicting which TL/manual grader is more accurate per image with respect to the RPs. This is the first work of its kind that considers relative agreeability/dis-agreeability between graders with respect to system level RPs as opposed to comparative assessment verses each other \cite{sanchez}\cite{Gopinath}\cite{GIRISH}. We find patterns in each vendor image stack that point towards the relative reliability of a particular grader ($G_1$ or $G_2$) for the specific image stack. In Table \ref{tab:compare}, we observe the aggregated DC per TL type, given the FSLs (Baseline, ParESN, U-net) are trained on that TL itself, i.e., FSL models trained on $G_1$ are tested on $G_1$ only and so on. We compare our observations to existing works that train on $G_1$ only. It is noteworthy that the goal of the FSL models in this work is to capture the relative ease/difficulty of annotation rather than perform automated segmentation and hence, the models here are trained on 5 images per image stack only. In Table \ref{tab:compare}, we find that there is a higher inter-observer variability in models when trained on different TLs (higher variability across TLs by the proposed models). Also, the ParESN model trained on $G_1 \cap G_2$ has the closest cyst segmentation performance when compared to deep learning models in \cite{sanchez}.

\begin{table}[ht]
   \centering
    \caption{Comparative Analysis of aggregated mean DC per grader with Existing Works.}\label{tab:compare} 
    \scalebox{0.95}
    {
    \begin{tabular}{|c|c c c|}\hline
    Method&$G_1$&$G_2$&$G_1 \cap G_2$\\ \hline
Proposed Baseline&	0.31&	0.35&	0.38\\
Proposed ParESN&	0.46&0.49&	0.51\\
Proposed U-net&	0.41&0.45&0.46\\ \hline
Oguz et. al. \cite{oguz}&	0.48&	0.48&	0.48\\
Esmarelli et. al. \cite{esmarelli}&	0.46&	0.45&	0.45\\
Venhuizen et al. \cite{sanchez}&0.56&	0.55&	0.54\\
Gopinath et. al. \cite{Gopinath}&0.67&	0.68&0.69\\\hline
 \end{tabular}}
    \vspace{-0.5cm}
\end{table}

Thus, the proposed system can automatically detect annotated image quality along with a confidence score metric to standardize the otherwise laborious manual quality assurance process. Future works can be directed towards active learning methodologies to sample the training images for the FSL models described here to further improve segmentation performances. Additionally, the proposed system can be analyzed on other medical and non-medical image stacks for automated quality assurance process.
\bibliographystyle{IEEEtran}
\bibliography{paper}

\ifCLASSOPTIONcaptionsoff
  \newpage
\fi

\end{document}


\title{Supplementary Material: FSL Framework to Reduce Inter-Observer Variability}
 \maketitle
 
 \IEEEpeerreviewmaketitle

 \section{The ParESN Model}
 
\begin{figure}[ht!]
    \centering
    \includegraphics[width=0.90\textwidth]{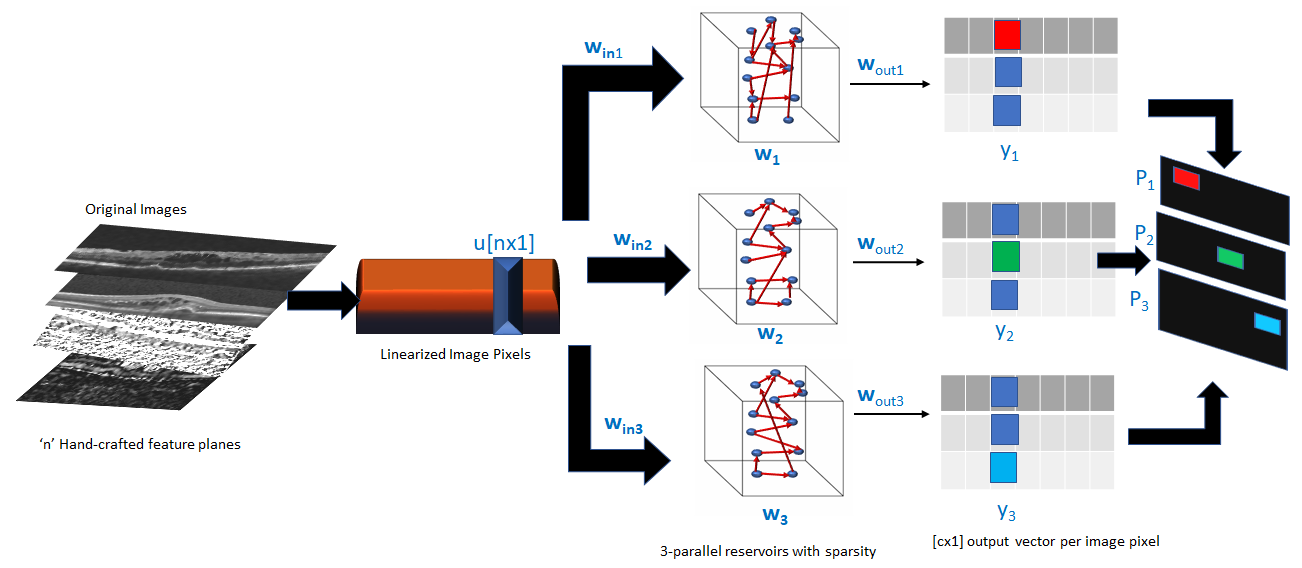}
    \caption{System setup of the proposed ParESN model. Each image and its 3 pre-processed planes are converted into input matrix $\mathbf{U}$ and fed to the 3 parallel ESN branches to update the reservoir state matrix $\mathbf{X}$ per branch. At the end of the training process, a $c$-dimensional vector is output per pixel location, where $c$ represents the number of classes to be predicted ($c=2$ for binary segmentation). The per-pixel output ($p(k) \in \{P_1, P_2, P_3$\}) is the class label with maximal probability. Each regional proposals (RPs), i.e. $P_1, P_2, P_3$ from the 3 parallel arms represent cyst-like pixels that are trained from the same images and similar training setups. To demonstrate qualitative overlap between the RPs, the RP from the top, middle and bottom ESN arms are visualized with the red, blue and green image planes respectively.}
    \label{esn}
\end{figure}

 The Parallel ESN Framework presented in this work is inspired by the previous works in \cite{sohiniesn} and \cite{esn2}. The primary difference with the model in \cite{sohiniesn} is the variation in importance for the hidden states with respect to a previous pixel in the same image vs. the same pixel in the previous image. With respect to \cite{esn2}, the parallel branches in this work generate similarly trained regional proposals (RPs) instead of a cumulative combination from the three parallel streams in \cite{esn2}. Also, the choice of three parallel layers in this work is optimal amongst the choice of $\{2, 3, 4, 5\}$ parallel RPs based on a grid search and leave-one-out cross validation across vendor image stacks. The detailed system setup is shown in Fig. \ref{esn}.
 The OCT cyst images per vendor stack represent a volumetric level scan. This implies if a large cyst appears in a scan, there is a high probability of the same cyst appearing in some shape and form in the previous and succeeding scans as well. The proposed setup is analogous to the video processing setup in \cite{sohiniesn}. Thus, the reservoir states per pixel location of subsequent images will be affected by the previous and current image, represented in \eqref{eq1}.

\begin{align}\label{eq1}
\mathbf{x_{\nu}}(k)=(1-\alpha) \,\mathbf{x_{\nu}}(k-1)\\ \nonumber
+\alpha \,f\big(\mathbf{W}_{\mathrm{in, \nu}} \,[1;\mathbf{u}(k)]+\mathbf{W_{\nu}}\,\mathbf{x_{\nu}}(k-1)\big).
\end{align}

At the end of the training stage, $W_{out,\nu}, \nu=\{1,2,3\}$ are computed for each parallel layer using \eqref{eq2}. 
\begin{align}\label{eq2}
\mathbf{w}_{\mathrm{out,\nu}}=\Big(\sum_{l=1}^{L}\mathbf{z}_{\nu,l}(k)\mathbf{z}^{\mathrm{T}}_{\nu,l}(k)+\lambda \, \mathbf{\mathds{1}}\Big)^{-1}\Big(\sum_{l=1}^{L}\mathbf{z}_{\nu,l}(k)y(k)\Big)
\end{align}
where, $\mathbf{z_{\nu,l}}(k)=[1;\mathbf{u}(k);\mathbf{x_{\nu}}(k)]$ are the extended system states for the 3 parallel layers, evaluated over $l=\{1,2 \dots L\}$ training images, and $y(k)$ represents the target label at pixel location $k$, and $\mathbf{\mathds{1}}$ represents identity matrix.

The leave-one-out cross validation experiment across vendor stacks helps identify the optimal parameter set $\{\alpha, \lambda\}$ in \eqref{eq1}, \eqref{eq2}, respectively. We observe $\alpha=[0.95]$ to be optimal from the search set of $[0.30:0.99]$ in increments of 0.02. This implies very low ``leaky-memory'' requirement for the data set \cite{DeepESN}. Also, the sensitivity to $\lambda$ is found to be very low, with $\lambda=10^{-5}$ being optimal for the setup in the search set of $[10^{-10}-0.1]$ in order increments of 10.

 \section{Qualitative Analysis of Target Label Selection Algotithm (TLSA)}
 The key contribution of our work is the use of the RPs to detect the ``best'' manual annotation at image level. Examples of the TLSA for TL vs noisy TL selection and for $G_1$ vs. $G_2$ selection are shown below.
 \subsection{Examples of TLSA against noisy labels}
\begin{figure}[ht]
	\centering
	\subfigure[Left: TL and noisy TLs. Right: RPs from ParESN model. Very small difference between the TL and noisy TLs (left) and high variability in regional overlap between RPs (right) requires manual intervention in all these images.]
	{\includegraphics[0.5\textwidth]{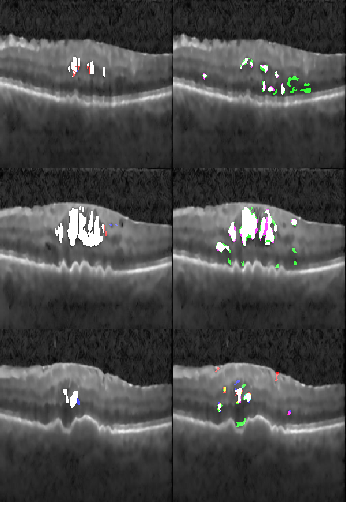}}
	\subfigure[Left: TL and noisy TLs. Right: RPs from ParESN model. In all these cases, a different RP is found to have the most overlap with TL or noisy TL, respectively. However, the mean overlap metric between the RPs and the TLs are the decisive metric to select the correct TL over its noisy counterparts in each case.]
	{\includegraphics[0.5\textwidth]{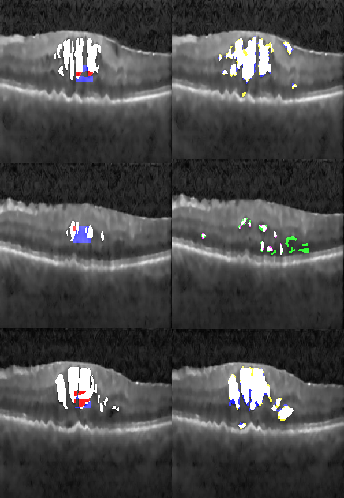}}
	\caption{TL vs. noisy TL assessment.}
	\label{noisy}
	\vspace{-0.5cm}
\end{figure}

 For this experiment the Random Crop and Paste function is invoked to generate noisy TLs. In Fig \ref{noisy}, the actual TLs are represented as red, noisy generations using RCAP as blue and their intersections as white regions, respectively. Also, the RPs $P_1,P_2, P_3$ are represented in red, green and blue planes, respectively. In Fig. \ref{noisy} qualitative explanations are provided for manual interventions and for automatic selection of actual TL over the noisy counterparts.
 
 \subsection{Examples of TLSA for best TL selection}
\begin{figure}[ht]
	\centering
	\subfigure[Left: TLs. Right: RPs from ParESN model. RPs are trained on $G_1$, but agree more with $G_2$ in blue.]
	{\includegraphics[0.5\textwidth]{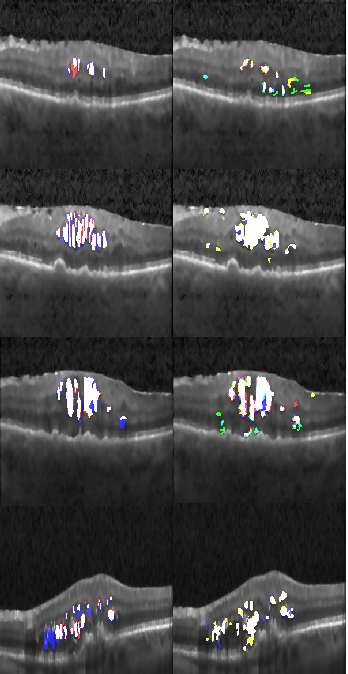}}
	\subfigure[Left: TLs. Right: RPs from ParESN model. RPs are trained on $G_2$, but agree more with $G_1$ in red.]
	{\includegraphics[0.5\textwidth]{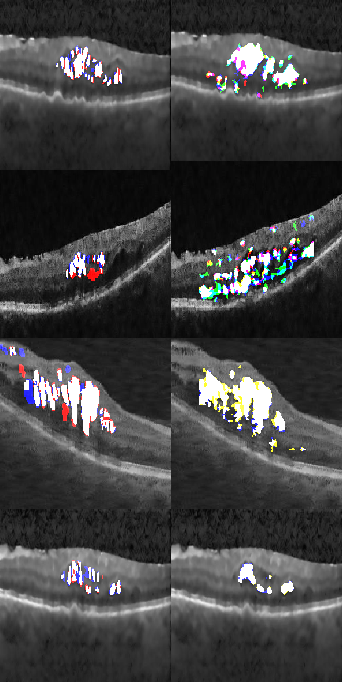}}
	\caption{Assessment of best TL selection.}
	\label{choice}
	\vspace{-0.5cm}
\end{figure}

 In Fig. \ref{choice}, qualitative examples of best TL selection are shown. In Fig. \ref{choice}, the left columns represent the TLs such that $G_1$ is in red, $G_2$ in blue and their intersection is in white. The right columns represent the RPs in red, green and blue planes respectively. In Fig. \ref{choice}(a), the Few-shot learning (FSL) models are trained on $G_1$, but the RPs agree more with $G_2$, and hence, $G_2$ is selected as the best label for all these images. In Fig. \ref{choice}(b), the FSL models are trained on $G_2$, but the RPs agree more with $G_1$, and hence $G_1$ is selected as the best label for all these images. The examples in Fig. \ref{choice} demonstrate the extent of variabilities by manual target labels $G_1$ and $G_2$. 
 
 For the images in Fig \ref{choice}(a), $G_1$ annotations represented many contiguous small cyst regions, based on which, RPs get trained to identify small contiguous cysts in test images. However, $G_1$ annotates some larger cyst areas in the test images as opposed to $G_2$ that detects smaller cysts. Thus, $G_2$ is preferable in all such examples in Fig \ref{choice}(a). For images in Fig. \ref{choice}(b), the RPs trained on $G_2$ have an affinity to detect large cysts that appear in $G_1$. Hence, in all these examples, $G_1$ is the preferred TL in spite of the FSL models being trained on $G_2$. 
 From the examples in Fig. \ref{choice}, we are able to qualitatively assess the importance of the TLSA for standardizing cyst segmentation to overcome inter-observer variabilities.

 \bibliographystyle{IEEEtran}
\bibliography{paper}